\documentclass[lettersize,journal]{IEEEtran}
\usepackage{amsmath,amsfonts}
\usepackage{algorithmic}
\usepackage{array}
\usepackage[caption=false,font=normalsize,labelfont=sf,textfont=sf]{subfig}
\usepackage{textcomp}
\usepackage{stfloats}
\usepackage{url}
\usepackage{verbatim}
\usepackage{graphicx}
\def\BibTeX{{\rm B\kern-.05em{\sc i\kern-.025em b}\kern-.08em
    T\kern-.1667em\lower.7ex\hbox{E}\kern-.125emX}}
\usepackage{balance}

\usepackage{multirow}
\usepackage{arydshln}
\usepackage{colortbl}

\usepackage{color}

\begin{document}

\title{Few-shot Learning for Cross-Target Stance Detection by Aggregating Multimodal Embeddings}

\author{
  Parisa Jamadi Khiabani, Arkaitz Zubiaga \\
  Queen Mary University of London, UK \\
  \{p.jamadikhiabani,a.zubiaga\}@qmul.ac.uk
}

\maketitle

\begin{abstract}
 Despite the increasing popularity of the stance detection task, existing approaches are predominantly limited to using the textual content of social media posts for the classification, overlooking the social nature of the task. The stance detection task becomes particularly challenging in cross-target classification scenarios, where even in few-shot training settings the model needs to predict the stance towards new targets for which the model has only seen few relevant samples during training. To address the cross-target stance detection in social media by leveraging the social nature of the task, we introduce CT-TN, a novel model that aggregates multimodal embeddings derived from both textual and network features of the data. We conduct experiments in a few-shot cross-target scenario on six different combinations of source-destination target pairs. By comparing CT-TN with state-of-the-art cross-target stance detection models, we demonstrate the effectiveness of our model by achieving average performance improvements ranging from 11\% to 21\% across different baseline models. Experiments with different numbers of shots show that CT-TN can outperform other models after seeing 300 instances of the destination target. Further, ablation experiments demonstrate the positive contribution of each of the components of CT-TN towards the final performance. We further analyse the network interactions between social media users, which reveal the potential of using social features for cross-target stance detection.
\end{abstract}

\section{Introduction}
\IEEEPARstart{I}{n} the information-driven world we now inhabit, a large amount of opinion texts can be found on the Web. The presence of this content is ever growing as social networking platforms become increasingly popular, which according to recent statistics are used by around 65\% of American adults \cite{perrin2015social}, attracting a great deal of public attention \cite{tian2020early}. However, given the volume of content posted daily, monitoring the opinions expressed in social media platforms remains a time-consuming task which is not manageable without the support of automated means \cite{liu2012sentiment,zubiaga2019mining}. Hence, there is a need to develop novel methods for automated classification and processing of these texts to determine the stance expressed in texts with the aim of mining public opinion. 

Stance classification is concerned with identifying a person’s or a post's standpoint towards a target \cite{biber1988adverbial}, which is generally classified as one of in favor of (supporting) or against (opposing) the target in question \cite{aldayel2021stance,jamadi2020improved,liu2010sentiment,biber1988adverbial,zubiaga2018longitudinal,mohammad2016semeval}. Stance classification from social media data is however a challenging task \cite{antoun2020arabert,zhang2020enhancing}, given the diverse and informal nature of social media data. Despite recent progress in stance classification, there is still substantial room for improvement, particularly when it comes to enabling generalisability of classifiers to deal with new targets \cite{li2021p}. This is the case in cross-target stance classification; where a classifier has seen training data associated with particular targets but needs to predict the stance towards new targets, of which the model has seen no or few instances. An ability to deal with new targets is however important, given the evolving nature of the targets intended for the analysis. For example, an interest for mining stances towards US President Donald Trump can eventually shift towards Joe Biden as the new president takes office.

Previous research in cross-target stance classification has introduced approaches that leverage the textual content of posts, generally by using transfer learning strategies \cite{wei2019modeling,zhang2020enhancing}. Previous research has however been limited to the sole use of the textual content of posts to determine the stance, hence overlooking the potential of other features inherent to social media. Following the intuition of the theory of homophily, which suggests that like-minded users will tend to follow each other and like each other's posts, we therefore propose further digging into network features for cross-target stance detection. Our objective here is both to demonstrate that network information can be uniquely valuable to enhance the cross-target stance detection task as well as to propose a novel methodology that effectively does so. We propose a novel method, CT-TN, which encapsulates text and network features through a proposed architecture that aggregates both feature types for improved stance classification. Our aim here, in line with much of the previous work, is to use a small number of instances from the destination target in the training phase through few-shot learning.

To the best of our knowledge, CT-TN is the first multimodal architecture for cross-target stance classification which combines text and network features. By experimenting on six different combinations of source and destination targets  in a few-shot cross-target stance detection scenario, we demonstrate the effectiveness of CT-TN to consistently outperform two state-of-the-art cross-target stance detection models as well as a state-of-the-art pre-trained language model, RoBERTa.

\textbf{Contributions.} Through our study, we make the following key contributions:

\begin{itemize} 
 \item We propose CT-TN (Cross-Target Text-Net) model, a model that encapsulates multimodal embeddings by integrating state-of-the-art text and graph embedding strategies for the cross-target stance classification task.

 \item We investigate the effectiveness of CT-TN in the few-shot cross-target stance detection task by using the P-Stance dataset, one of the largest stance datasets which enables experimenting with combinations of six different source and destination target pairs.
 
 \item We perform ablation experiments to investigate the impact of the different components that form CT-TN, assessing whether and the extent to which each of them is contributing positively to improved performance of the model.
 
 \item In addition to our initial experiments with 400 shots from the destination target used for training, we further investigate scenarios where fewer shots are available, investigating its impact on the CT-TN model, and assessing how many shots the model needs to perform competitively.
\end{itemize}

We find that our model can consistently outperform state-of-the-art text-based cross-target stance detection models such as TGA-Net and CrossNet. Ablation experiments with CT-TN alternatives demonstrates the contribution of its different components. While we demonstrate the effectiveness of CT-TN, we also observe that it becomes less effective when we dramatically reduce the number of shots used during training, suggesting that the contribution of carefully integrated network features becomes useful after 300+ shots, but is less reliable when fewer shots (100 or 200) are available.

\textbf{Paper structure.} The remainder of this paper is organised as follows: Section \ref{sec:related-work} discusses work related to ours looking at the challenges of cross-target stance detection as well as the use of multimodal embeddings for stance detection. We introduce our proposed method CT-TN in Section \ref{sec:Methodology}, followed by the experiment settings which we describe in Section \ref{sec:Experiments}. We present our experiments results in Section \ref{sec:Results} as well as we further discuss and delve into them in Section \ref{sec:Discussion}. We conclude the paper in Section \ref{sec:conclusion}.

\section{Related Work}
\label{sec:related-work}

We discuss related work in two main directions: research on cross-target stance detection and the use of multimodal embeddings in stance detection.

\subsection{Cross-target Stance Detection}
\label{sec:Cross-target Stance Detection}

Despite a substantial body of research in stance detection in recent years \cite{kuccuk2020stance,aldayel2021stance,cao2022stance}, the more challenging task of cross-target stance detection has received less attention. One of the first approaches to cross-target stance detection is \textbf{Bicond} \cite{augenstein2016stance}, which combined multiple layers of LSTM models in different settings encoding the texts from left to right and from right to left. \cite{xu2018cross} developed \textbf{CrossNet}, which added an Aspect Attention Layer to the Bicond model, which enabled discovering domain-specific aspects for cross-target stance inference, utilising self-attention to signal the core parts of a stance-bearing sentence. Their model consists of four main layers: Embedding Layer, Context Encoding Layer, Aspect Attention Layer, and Prediction Layer. Their model showed to outperform the Bicond model.

A different type of approach focused on transferable topic modelling. This is the case of the \textbf{VTN} model \cite{wei2019modeling}. This model uses shared latent topics between two targets as transferable knowledge in order to achieve model adaptation. The latent topics are determined by using Neural variational inference \cite{miao2016neural}. Another model by \cite{zhang2020enhancing}, called \textbf{SEKT}, proposed to leverage external knowledge to perform stance detection across targets. Still limited to processing textual content, they proposed to generate a semantic-emotion heterogeneous graph (SE-graph) which is fed to a GCN and a BiLSTM for the classification.

One of the best-known models among those presented recently is Topic-Grouped Attention (\textbf{TGA}), introduced by \cite{allaway2020zero}, and is one of our key baseline models. The model consists of four main phases: (i) Contextual Conditional Encoding, i.e. using contextual emebddings like BERT to embed a document and topic together, (ii) Generalized Topic Representations (GTR), i.e. using Ward hierarchical clustering, (iii) Topic-Grouped Attention, i.e. using learned scaled dotproduct attention (compute similarity scores), and (iv) Label Prediction, i.e. feed-forward neural network  to compute the output probabilities. This model proved competitive in comparison with a range of other baseline models, showing greater generalisability across different targets than other models.

However, existing approaches are limited to processing the textual part of the posts only for the cross-target stance detection. Here, we further design this research by proposing the first model that leverages network features in addition to text, through our proposed model CT-TN. To the best of our knowledge, this is the first study that takes advantage of using network/social information along with text into the cross-target stance detection task. In our experiments, in addition to CT-TN, we also experiment with TGA-Net and CrossNet as competitive baseline models.

\subsection{Multimodal Embeddings and Stance Detection}

\textbf{Text and Graph Embeddings.} There has been a substantial body of research in recent years in developing embedding approaches that deal with either texts or graphs separately.

When it comes to \textbf{text-based embeddings}, non-contextualised representations such as Word2vec \cite{mikolov2013efficient} and GloVe \cite{pennington2014glove} were soon followed by more sophisticated, contextualised representations such as ELMo \cite{peters1802deep} and OpenAI's GPT \cite{radford2018improving}. The latter use unidirectional language models in order to learn general language representations, restricting the efficiency of the pre-trained models.

Recently, researchers spend more time on applying transfer learning by ﬁne-tuning large pre-trained language models for downstream NLP/NLU tasks, including a small number of examples which achieves distinguish performance improvement regarding these tasks. Despite the fact that pre-trained language models are used for this approach, they suffer from a main limitation which is needing huge corpora for pre-training plus the high computational cost of days needed for training \cite{antoun2020arabert}. 

Transformer models \cite{vaswani2017attention}, including the likes of BERT \cite{devlin2018bert} have more recently become the state-of-the-art models for text representation in text classification. Models under this category include BERT, RoBERTa \cite{liu2019roberta}, XLM \cite{lample2019cross} and XLM-R \cite{conneau2019unsupervised}. In our work, we make use of the RoBERTa model as a component of CT-TN, as well as a baseline model when used on its own.

When it comes to \textbf{graph embeddings}, different approaches have been proposed which operate at the node, sub-graph or different levels of granularity. These types of model include unsupervised techniques like DeepWalk \cite{perozzi2014deepwalk}, a method based on random walks, utilizing local neighborhoods of a node. A more popular method for generating embeddings from graphs is Node2Vec \cite{grover2016node2vec}, which consists of a flexible biased random walk procedure to explore networks, being one of the first Deep Learning attempts to learn from graph data. There are similarities between Deepwalk \cite{perozzi2014deepwalk} and Node2Vec \cite{grover2016node2vec} in that they both maximise the probability of node co-occurrences in sampled random-walks approaches across the graph. However there is a difference based on how the random walks are sampled. The former uses unbiased random walks, whereas the latter biases the random walks using two random walk hyperparameters return parameter (p) and in-out parameter (q). The p parameter controls the likelihood of immediately revisiting a vertex; and the q parameter q is responsible to controlling the likelihood that the walk revisits a vertex’s one-hop neighbourhood. Deepwalk can in fact be viewed as a special case of node2vec where p=q=1. Another line of research towards graph embedding methods focuses on supervised techniques that utilise custom neural network components that can operate directly on graphs. One of the methods in this line of research is that of the wave architecture \cite{matlock2019deep}, which has the advantage of propagating long-range information across a graph compared to the graph convolution architectures. In our case, however, our focus in this work is on unsupervised graph embedding methods, and use of wave based architectures and other supervised strategies is left for future work.

The key focus of our work is on leveraging a cross-target stance architecture that can effectively combine textual and network information together in comparison to state-of-the-art baseline models that focus solely on textual content. Hence, we leverage existing graph embedding models rather than building new architectures, with the aim of directly tackling our key objective. To this aim, we use PecanPy \cite{liu2021pecanpy} as the component to implement graph embeddings, which is an optimised implementation of Node2Vec that makes it more efficient thanks to its paralellisation. We use PecanPy to extract embeddings from three different types of network information: followers, friends and likes. 

\textbf{Multimodal Embeddings in Stance Detection.} Combining textual embeddings with graph embeddings has been studied in previous research, however barely in the context of stance detection. This is the case of \cite{ostendorff2019enriching}, who proposed an approach to enrich a BERT transformer by incorporating knowledge graph embeddings trained from Wikidata. Through experiments on a book classification task, they showed that their method could outperform other text-only baseline methods.

Despite the inherently social nature of the stance detection task, the vast majority of the research has been limited to textual features and embeddings. The main exception to this is the work by \cite{aldayel2021stance}, who demonstrated that social signals could also be helpful to predict the stance expressed by a user, suggesting that it could even be possible to predict the stance of a user who has not posted anything, solely based on their network. This work is however limited to in-target stance detection.

To the best of our knowledge, no work has studied the multimodal aggregation of textual and graph embeddings for cross-target stance detection. Through the introduction of CT-TN, we aim to propose the first approach that effectively achieves this in the cross-target stance detection task.

\section{Methodology}
\label{sec:Methodology}

\subsection{Problem Formulation}
\label{ssec:formulation}

We formulate the stance detection task as that in which each of the posts in a collection $P = \{p_1, p_2, ..., p_n\}$ has to be classified into one of the stances $S = \{favor, against\}$, where each post $p_i$ expresses a stance towards a target $t$. We have a dataset where each post expresses a stance towards one of the targets in a collection $T = \{t_1, t_2, ..., t_m\}$. The cross-target stance detection task consists in predicting the stance expressed in posts referring to target $t_i$, where the training data is composed of posts referring to other targets excluding $t_i$, hence requiring a transfer of knowledge from a set of targets to another. In the few-shot cross-target stance detection scenario, however, we experiment with a small number of instances referring to $t_i$ incorporated into the training data, to relax this cross-target scenario. In our particular case, we experiment with 400 instances (i.e. 400 shots) pertaining to $t_i$ incorporated into the training data in the few-shot scenario; in subsequent experiments, we test with smaller numbers of instances in the few-shot scenario, namely 100, 200 and 300.

\subsection{Proposed Method: CT-TN}
\label{ssec:cttn}

\begin{figure}[htb]
	\centering
		\includegraphics[width=0.5\textwidth]{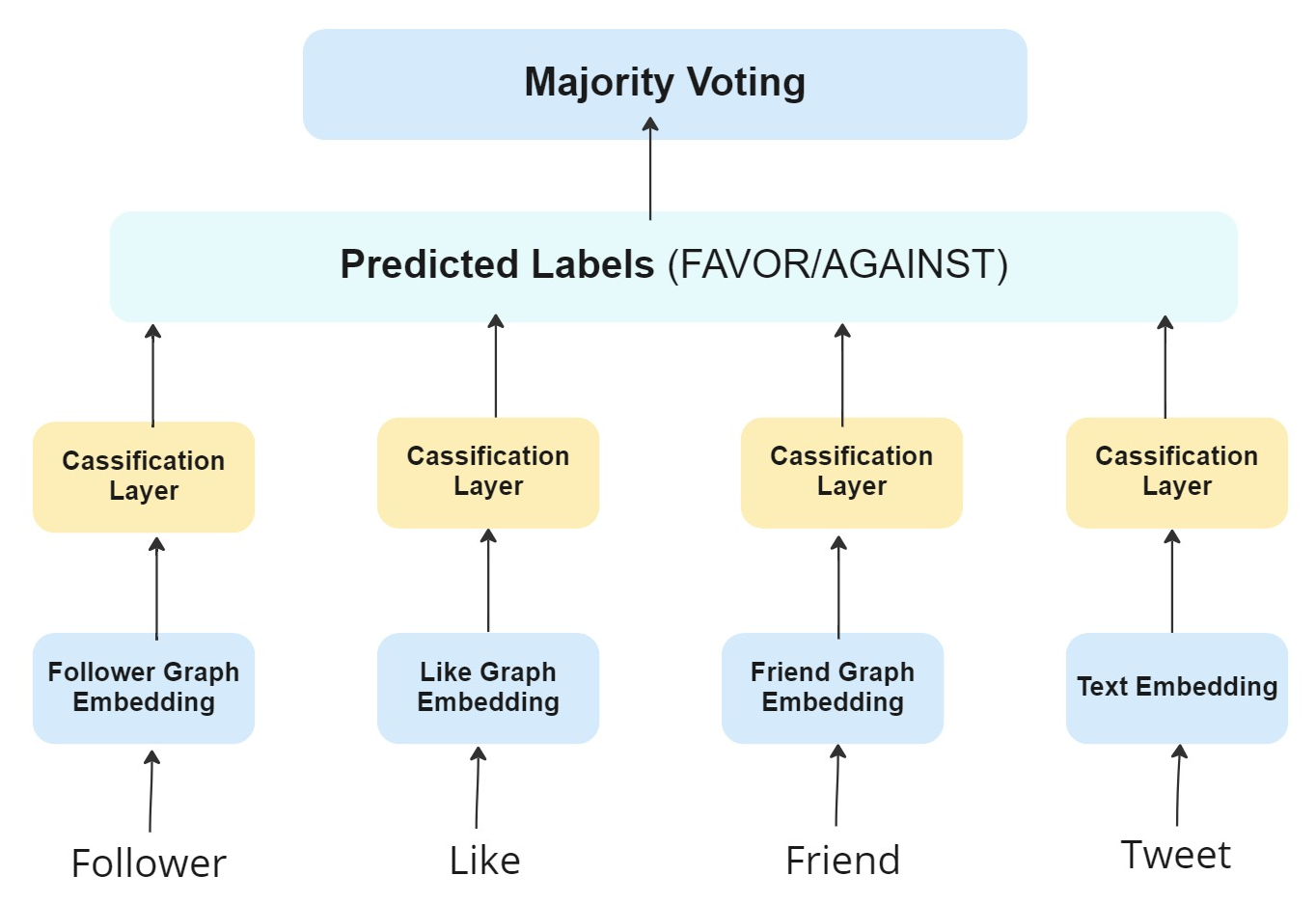}
	\caption{Architecture of the proposed model, CT-TN.}
	\label{fig:architecture}
\end{figure}

The CT-TN architecture consists of three main types of components: (i) text-based embedding generation and classification, (ii) three instances of network encoding components for graph-based embedding generation and classification, which are used for feeding followers, friends and likes, and (iii) output aggregation. The first two types of components are executed in parallel to produce isolated stance predictions, after which the final component aggregates the predictions for final output. Figure \ref{fig:architecture} demonstrates general architecture of our proposed model. In what follows we describe the specifics of these three components.

\textbf{Component \#1: Text-based classification: Contextual Conditional Encoding.} This component takes in the textual content of the input posts using bidirectional conditional encoding (conditioning the document representation on the topic) layer followed by a feed-forward neural network. Previous works have shown the advantage of utilising contextual embeddings \cite{devlin2018bert}. We embed the user generated text through the RoBERTa language model \cite{liu2019roberta} to embed a document and topic jointly (768 dimension vector). RoBERTa can take as input either one or two sentences, and uses the special token [SEP] to separate them. In order to input both the textual content and target information associated with the post, we feed the following to the model: ``[CLS] + target + [SEP] + context”. This component produces an output with its own prediction for the stance of a particular post, as either supporting or opposing.

\textbf{Component \#2: Graph-based classification: Network Encoding.} The CT-TN model uses three different instances of the network encoding model for graph-based classification, for representing three types of inputs: followers, friends and likes. To generate embeddings using the Node2Vec architecture \cite{grover2016node2vec}, we use the PecanPy implementation \cite{liu2021pecanpy} which optimises its performance. The node embeddings calculated using PecanPy (128 dimension vector) can be used as feature vectors in a downstream task such as node classification. In our case, user IDs are considered as graph nodes and the relationships between users (friends/followers/likes) are provided as graph edges. Each of the three components implemented here through network encodings produce their own predictions on a particular post (supporting or opposing).

\textbf{Component \#3: Output aggregation.} The final component takes as input the predictions made by all the above components, i.e. the text-based and three network-based components. To aggregate the predictions of all these four components, the ``output aggregation'' component implements a voting ensemble (or a ``majority voting“ strategy) which combines the different predictions, ultimately choose the class with a larger number of votes. While we did test alternative methods for combining different embedding inputs, such as concatenation of embeddings, both with and without normalisation of the input embeddings, the `majority voting' strategy proved clearly more effective in initial experiments and hence was the option we finally selected to develop CT-TN.

\subsection{Model Hyperparameters}

We use the RoBERTa base model (roberta-base-cased) as our pre-trained language model, due to its improved performance over similar transformer models such as BERT \cite{adoma2020comparative}. It consists of 12 transformer layers, each of which adopts a hidden state size of 768 and 12 attention headers. Training for RoBERTa text embedding is performed with batch size b = 128, dropout probability d = 0.2, learning rate= 3e-5 (AdamW optimiser) and 40 training epochs. While we trained graph embedding models as follow: batch size b = 128, dropout probability d = 0.2, learning rate= 1e-2 (SGD optimiser) and 100 training epochs. For model training, we use multi-class cross-entropy loss.

While previous research has addressed the stance detection task in both 3-class \cite{xu2016overview,mohammad2016dataset} and 2-class \cite{murakami2010support,abbott2016internet} settings, our focus here is on the latter, while the extension of our proposed model to 3-class is beyond the scope of this paper.

\section{Experiments}
\label{sec:Experiments}

Next, we provide the details of the dataset we use in our research, as well as the baseline methods we compare our method against, followed by experiment settings and evaluation metrics used in our work.

\subsection{Dataset}

We chose to use the P-Stance dataset \cite{li2021p}, given that it is an order of magnitude larger than other publicly available datasets and because it provides more than one target, as required for our research in cross-target stance detection. The P-Stance dataset is originally composed of tweets pertaining to three different political figures as targets: ``Donald Trump,” ``Joe Biden,” and ``Bernie Sanders.”  A sample of the P-Stance dataset is provided as Table \ref{tab:pstance}.

\begin{table*}[htb]
 \normalsize
 \begin{center}
  \caption{A sample of the P-Stance dataset.}\label{tab:pstance}
  \begin{tabular}{|p{12cm}|l|l|}
   \hline
   Tweet  & Target    & Stance  \\
   \hline
   How Joe   Biden would make community college free and fix student loans via @politico & Joe Biden & FAVOR \\
   \hline
   Glad our   GREAT President called out the so called whistleblower. If there is a Senate   trail, they may call the whistleblower to testify. BTW Trump is not impeached   until crazy Nancy send the articles over to the Senate. Trump will not be   convicted. Vote Trump 2020 & Donald   Trump   & FAVOR   \\
   \hline
   \#Bernie   Sanders says he's 'one of the poorer members of the \#UnitedStatesSenate'   \#BetOil is A Multimillionaire,\#Warren has A 5 Million dollar Home,\#Hillary   HAS several Mansions plus A Super Millionaire! Whats Your Point?                                          & Bernie   Sanders & AGAINST \\
   \hline
  \end{tabular}
 \end{center}
\end{table*}

The original P-Stance dataset, as published by the authors, only contains the tweet texts associated with their stance labels. This original dataset lacked the network information that we needed. Upon request, the authors kindly shared 9,307 tweet IDs, which we use to reconstruct and expand the dataset. This includes retrieving full tweet metadata, from which we can extract the user IDs, which would then allow us to retrieve the user network.

\begin{figure}[htb]
	\centering
		\includegraphics[scale=.6]{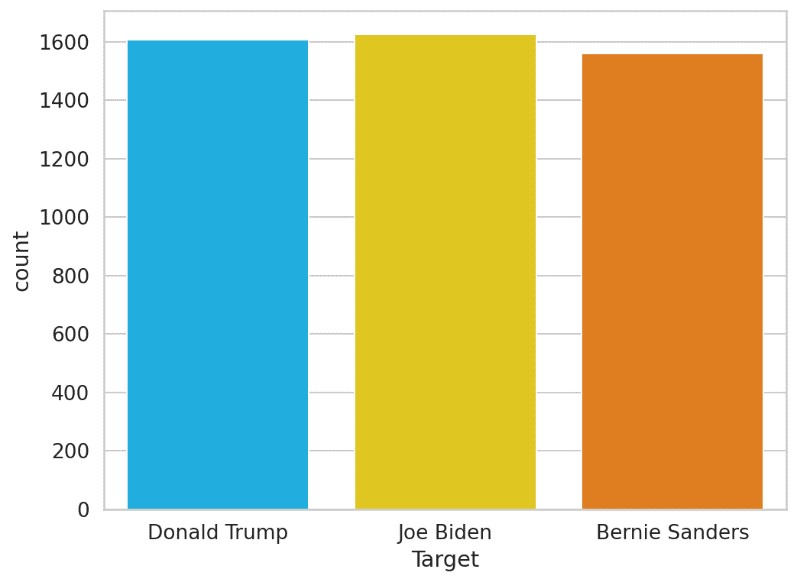}
	\caption{Distribution of targets in the P-Stance dataset.}
	\label{fig:targets}
\end{figure}

Given the focus of our research in 2-class classification (i.e. favor or against), we retrieve metadata for the tweets associated with these categories. This led to 4,212 tweets with available metadata. The resulting dataset has a distribution of labels as shown in Table \ref{tab:pstance-stats}, and a distribution in the number of tweets per target as shown in Figure \ref{fig:targets}. While the number of tweets across targets is very similar, we observe some differences in the number of favor and against tweets, with Donald Trump having the largest ratio of against tweets, and Bernie Sanders having the largest ratio of favor tweets. For these available tweets, we further complement the data retrieval as described below.

\begin{table}[htb]
 \normalsize
 \begin{center}
  \caption{The statistics of the resulting dataset.}\label{tab:pstance-stats}
  \begin{tabular}{|p{2.5cm}|c|c|c|}
   \hline
   \textbf{Target}  & \textbf{Favor} & \textbf{Against}        & \textbf{Avg. length}  \\
   \hline
   Donald Trump &  519 & 947      & 34.7   \\
   \hline
   Joe Biden & 702 & 716   & 33.7   \\
   \hline
   Bernie Sanders  & 776 & 553 & 31.5 \\
   \hline
  \end{tabular}
 \end{center}
\end{table}

Having the collection of tweet IDs, tweet metadata and user IDs, we proceeded with the retrieval of additional data including networks of the users (followers and friends) and likes (tweets they liked from other users). We detail each of these additional data collection steps next:

\begin{itemize}
 \item \textbf{Retrieval of followers:} Followers include the set of users who follow a particular user. For the user IDs in the dataset, we retrieve the complete list of followers for each user. This leads to a list of user IDs followed by each user, which allows us to build a network of followers.

 \item \textbf{Retrieval of friends:} Friends constitute the set of users followed by a user. Similar to the list of followers, this provides a list of user IDs per user, with which we can build a network.

 \item \textbf{Retrieval of likes:} For each user, we retrieve the tweets they liked from others. Given that we are interested in building networks of users, we obtain the user IDs pertaining to the tweets liked by the users. This again allows us to build a network, very similar to the friend / follower networks above, in this case based on the user IDs whose tweets have been liked by each user.
\end{itemize}

Hence, for each user, we have four features: (i) the textual content of their tweet, (ii) the network of followers, (iii) the network of friends, and (iv) the network of likes. Each of these is associated with a favor or against label, which we aim to predict.

After aggregating all these four features, we end up with a dataset of 4,144 tweets (posted by 3,871 distinct users) for which we have all features available.

\subsection{Baseline Methods}

We evaluate and compare our model with several strong baselines, including two of the main state-of-the-art cross-target stance detection models as well as the widely-used Transformer model RoBERTa:

\begin{itemize}
 \item \textbf{CrossNet \cite{xu2018cross}:} This model is a variant of BiCond, which leverages a self-attention layer to capture important words in the input text.
 
 \item \textbf{TGA-NET \cite{allaway2020zero}:} A new model has been proposed for (few-shot) cross-target stance detection that implicitly captures relationships between topics using generalized topic representations.
 
 \item \textbf{RoBERTa \cite{liu2019roberta}:} The method fine-tunes a pre-trained BERT model to perform cross-target detection. Specifically, we convert the given context and target to “[CLS] + target + [SEP] + context” structure for source and target domain, respectively.
\end{itemize}

Note that all of the above baselines make use of the textual content of the posts, as opposed to our proposed CT-TN also incorporating network information.

\subsection{Experiment Settings}

Experiments for the proposed few-shot cross-target approach are conducted in 100-shot, 200-shot, 300-shot, and 400-shot settings (e.g. injecting N samples of destination target into (source-target based) train data and then predicting the stance on the test data consist of only destination target) with 5 different random seeds: 24, 524, 1024, 1524, and 2024. Then we average the 5 seeds’ results per shot.

\subsection{Evaluation Metrics}

In line with previous research in stance detection \cite{augenstein2016stance,xu2018cross, allaway2020zero}, we also adopt the macro-averaged F1 score ($MacF_{avg}$) as the main metric to evaluate the performance in our experiments. In our case with binary classification involving the support and oppose classes, the resulting metric is the arithmetic mean of the F1 scores for each class, as follows:

\begin{equation}
 MacF_{avg} = \frac{F1_{support} + F1_{oppose}}{2}
\end{equation}

where each of $F1_{support}$ and $F1_{oppose}$ is defined as follows:

\begin{equation}
 F1_c = \frac{2 * precision_c * recall_c}{precision_c + recall_c}
\end{equation}

\section{Results}
\label{sec:Results}

We next present results of our proposed CT-TN model. We first discuss results of the model compared to a set of competitive baselines. We then delve into the results by analysing the performance of ablated versions of the model, and by looking at the impact of the number of shots used for training.

\subsection{CT-TN vs Baselines}

\begin{table*}[htb]
 \normalsize
 \begin{center}
  \caption{Macro-averaged F1 scores for CT-TN vs baseline models.}
  \label{tab:cttn-vs-baselines}
  \begin{tabular}{|l||cc|cc|cc||c|}
   \cline{2-8}
   \multicolumn{1}{l|}{Source} & Trump   & Sanders & Sanders & Biden   & Trump   & Biden & \multirow{2}{*}{Average} \\
   \multicolumn{1}{l|}{Test}   & Sanders & Trump   & Biden   & Sanders & Biden   & Trump &                          \\
   \hline
   RoBERTa   & 0.53 & 0.59 & \textbf{0.78} & 0.66 & \textbf{0.77} & 0.62 & 0.66 \\
   CrossNet  & 0.46 & 0.51 & 0.69 & 0.58 & 0.6  & 0.54 & 0.56 \\
   TGA-Net   & 0.57 & 0.6 & 0.69 & 0.6 & 0.69 & 0.59 & 0.62 \\
   \hdashline
   CT-TN     & \textbf{0.72} & \textbf{0.8}  & \textbf{0.78} & \textbf{0.73} & \textbf{0.77} & \textbf{0.82} & \textbf{0.77} \\
   \hline
  \end{tabular}
 \end{center}
\end{table*}

Table \ref{tab:cttn-vs-baselines} shows the results of CT-TN for the six combinations of source-destination targets under consideration, compared with the baseline models RoBERTa, CrossNet and TGA-Net. In addition to the results for each of the target pairs, we also show the average performance of each model across all pairs.

We observe that CT-TN consistently outperforms both cross-target stance detection models, CrossNet and TGA-Net, when we look at each target pair independently as well as at the overall average absolute improvements of 0.21 and 0.15, respectively.

CT-TN also performs remarkably better for than RoBERTa for a number of target pairs, not least Trump-Sanders, Sanders-Trump and Biden-Trump, with absolute improvements of 19\%, 21\% and 20\% respectively. This improvement is more modest for Biden-Sanders (7\%), with similar performances for the Sanders-Biden and Trump-Biden (0\%) target-pairs. On average, CT-TN still outperforms RoBERTa by 0.11, showing that it is more consistent across targets and overall more reliable. We believe that the strongest improvements of CT-TN with respect to the baselines come particularly for targets with significantly different ideology (i.e. those combining Trump and Sanders, and Trump and Biden); this suggests that for more distant targets, textual models may be more limited in capturing these substantial linguistic differences, whereas a network-based model generalises better in these situations.

\subsection{Analysis of Results per Class}

We next delve into the performance scores of the models broken down by category: favor and against. Table \ref{tab:cttn-vs-baselines-by-class} shows the results for the favor and against categories. We see that the improvement of the CT-TN model with respect to the baselines is consistent across both classes, hence showing that CT-TN provides a positive boost that impacts both classes positively. The extent of the improvement across classes is also consistent with the overall results shown above, as we see that the set of target pairs where CT-TN achieves the highest improvement matches those with the highest improvement in the overall results, i.e. Trump-Sanders, Sanders-Trump and Biden-Trump. Overall, CT-TN achieves improvements of 12\% in both the favor and against class over the second best model, RoBERTa.

\begin{table}[htbp]
\caption{Confusion matrices of CT-TN model vs. RoBERTa model for six target pairs}
\addtolength{\tabcolsep}{-1pt}
\normalsize
\begin{tabular}{|c|c|c|c|c|c|c|}
\cline{1-3}
\cline{5-7}
\multicolumn{3}{c}{\textbf{Biden-Sanders}} &       & \multicolumn{3}{c}{\textbf{Sanders-Biden}} \\
\cline{1-3}
\cline{5-7}
\multicolumn{3}{c}{\textit{CT-TN}} &    & \multicolumn{3}{c}{\textit{CT-TN}} \\
\cline{1-3}
\cline{5-7}
  & Against       & Favor &  &    & Against       & Favor \\
Against & 133           & 47    &  & Against & 123           & 50    \\
Favor   & 63            & 156   &  & Favor   & 47            & 179   \\
\cline{1-3}
\cline{5-7}
\multicolumn{3}{c}{\textit{RoBERTa}} &    & \multicolumn{3}{c}{\textit{RoBERTa}} \\
\cline{1-3}
\cline{5-7}
  & Against       & Favor &  &    & Against       & Favor \\
Against & 58            & 122   &  & Against & 53            & 120   \\
Favor   & 27            & 192   &  & Favor   & 51            & 175   \\
\cline{1-3}
\cline{5-7}
& & & & & & \\
\cline{1-3}
\cline{5-7}
\multicolumn{3}{c}{\textbf{Biden-Trump}}  &       & \multicolumn{3}{c}{\textbf{Trump-Biden}} \\
\cline{1-3}
\cline{5-7}
\multicolumn{3}{c}{\textit{CT-TN}} &    & \multicolumn{3}{c}{\textit{CT-TN}} \\
\cline{1-3}
\cline{5-7}
  & Against       & Favor &  &    & Against       & Favor \\
Against & 208           & 9     &  & Against & 180           & 59    \\
Favor   & 66            & 116   &  & Favor   & 34            & 126   \\
\cline{1-3}
\cline{5-7}
\multicolumn{3}{c}{\textit{RoBERTa}} &    & \multicolumn{3}{c}{\textit{RoBERTa}} \\
\cline{1-3}
\cline{5-7}
  & Against       & Favor &  &    & Against       & Favor \\
Against & 207           & 10    &  & Against & 183           & 56    \\
Favor   & 132           & 50    &  & Favor   & 43            & 117   \\
\cline{1-3}
\cline{5-7}
& & & & & & \\
\cline{1-3}
\cline{5-7}
\multicolumn{3}{c}{\textbf{Sanders-Trump}} &       & \multicolumn{3}{c}{\textbf{Trump-Sanders}} \\
\cline{1-3}
\cline{5-7}
\multicolumn{3}{c}{\textit{CT-TN}} &    & \multicolumn{3}{c}{\textit{CT-TN}} \\
\cline{1-3}
\cline{5-7}
  & Against       & Favor &  &    & Against       & Favor \\
Against & 179           & 10    &  & Against & 162           & 57    \\
Favor   & 71            & 139   &  & Favor   & 59            & 121   \\
\cline{1-3}
\cline{5-7}
\multicolumn{3}{c}{\textit{RoBERTa}} &    & \multicolumn{3}{c}{\textit{RoBERTa}} \\
\cline{1-3}
\cline{5-7}
  & Against       & Favor &  &    & Against       & Favor \\
Against & 147           & 42    &  & Against & 44            & 175   \\
Favor   & 149           & 61    &  & Favor   & 36            & 144   \\
\cline{1-3}
\cline{5-7}
\label{tab:confusion-matrices}
\end{tabular}
\end{table}

To further delve into the predictions broken down by class and to assess the consistent effectiveness of CT-TN across classes, we look at the confusion matrices of predictions. In the interest of a focused analysis, we present confusion matrices for the text-only baseline model (RoBERTa) as well as our proposed CT-TN model that combines graph and textual embeddings (see Table \ref{tab:confusion-matrices}. Looking at the pairwise comparison of predictions of CT-TN and RoBERTa for each of the source-destination target pairs, we observe that CT-TN is generally better at making more accurate predictions for both the favor and against classes. Exceptions include the favor class in the Biden-Sanders pair, the against class in the Trump-Biden pair and the favor class in the Trump-Sanders pair. CT-TN is consistently better in all the other cases. A look at the aggregate confusion matrix combining all target pairs (see Table \ref{tab:agg-confusion-matrix}) shows that CT-TN is clearly better overall in making predictions for both classes, with 985 correct `against' predictions (vs 692 by RoBERTa) and 837 correct `favor' predictions (vs 739 by RoBERTa).

\begin{table}[htb]
 \centering
 \normalsize
 \begin{tabular}{c|cc}
  \hline
  \multicolumn{3}{c}{\textbf{CT-TN}} \\
  \hline
    & Against & Favor \\
  Against & 985 & 232 \\
  Favor & 340 & 837 \\
  \hline
  & & \\
  \hline
  \multicolumn{3}{c}{\textbf{RoBERTa}} \\
  \hline
    & Against & Favor \\
  Against & 692 & 525 \\
  Favor & 438 & 739 \\
  \hline
 \end{tabular}
 \caption{Aggregate confusion matrix combining predictions across all six target pairs.}
 \label{tab:agg-confusion-matrix}
\end{table}

\begin{table*}[htb]
 \normalsize
 \begin{center}
  \caption{Macro-averaged F1 scores for the FAVOR and AGAINST classes with CT-TN vs baseline models.}
  \label{tab:cttn-vs-baselines-by-class}
  \begin{tabular}{|l||cc|cc|cc||c|}
   \cline{2-8}
   \multicolumn{1}{l|}{Source} & Trump   & Sanders & Sanders & Biden   & Trump   & Biden & \multirow{2}{*}{Average} \\
   \multicolumn{1}{l|}{Test}   & Sanders & Trump   & Biden   & Sanders & Biden   & Trump &                          \\
   \hline
   \multicolumn{8}{c}{`Favor' class} \\
   \hline
   RoBERTa   & 0.64 & 0.5 & 0.78 & 0.72 & 0.72 & 0.47 & 0.64 \\
   CrossNet  & 0.4 & 0.47 & 0.68 & 0.54 & 0.6  & 0.49 & 0.53 \\
   TGA-Net   & 0.55 & 0.56 & 0.7 & 0.67 & 0.65 & 0.47 & 0.6 \\
   \hdashline
   CT-TN     & \textbf{0.69} & \textbf{0.78}  & \textbf{0.79} & \textbf{0.74} & \textbf{0.75} & \textbf{0.78} & \textbf{0.76} \\
   \hline
   \multicolumn{8}{c}{`Against' class} \\
   \hline
   RoBERTa   & 0.44 & 0.68 & \textbf{0.77} & 0.59 & \textbf{0.8} & 0.75 & 0.67 \\
   CrossNet  & 0.5 & 0.55 & 0.7 & 0.61 & 0.63  & 0.6 & 0.6 \\
   TGA-Net   & 0.59 & 0.61 & 0.65 & 0.54 & 0.73 & 0.7 & 0.64 \\
   \hdashline
   CT-TN     & \textbf{0.76} & \textbf{0.81}  & 0.76 & \textbf{0.75} & \textbf{0.8} & \textbf{0.84} & \textbf{0.79} \\
   \hline
  \end{tabular}
 \end{center}
\end{table*}

\subsection{Ablated versions of CT-TN}

\begin{table*}[htb]
 \normalsize
 \begin{center}
  \caption{Macro-averaged F1 on full CT-TN vs ablated versions of CT-TN.\\Li: like, Fr: friends, Fl: followers, Rb: RoBERTa.}
  \label{tab:cttn-vs-ablated}

  \begin{tabular}{|p{0.5cm}p{0.5cm}p{0.5cm}p{0.5cm}||cc|cc|cc||c|}
   \cline{5-11}
   \multicolumn{4}{r|}{Source} & Trump   & Sanders & Sanders & Biden   & Trump   & Biden & \multirow{2}{*}{Average} \\
   \multicolumn{4}{r|}{Test}   & Sanders & Trump   & Biden   & Sanders & Biden   & Trump &                          \\
   \hline
   Li & Fr & Fl & Rb & & & & & & & \\
   \hline
     &   &   & x & 0.64 & 0.5 & \textbf{0.78} & 0.72 & 0.72 & 0.47 & 0.64 \\
   \hdashline
   x &   &   &   & 0.71 & \textbf{0.83} & 0.74 & \textbf{0.73} & 0.74 & \textbf{0.82} & 0.76 \\
	 & x &   &   & 0.7  & 0.79 & 0.75 & 0.71 & 0.75 & 0.81 & 0.75 \\
     &   & x &   & 0.69 & 0.8  & 0.74 & 0.7  & 0.74 & 0.8  & 0.75 \\
   \hdashline
   x &   &   & x & 0.63 & 0.79 & 0.77 & 0.7  & 0.76 & 0.74 & 0.73 \\
	 & x &   & x & 0.63 & 0.71 & 0.76 & 0.69 & \textbf{0.77} & 0.73 & 0.72 \\
     &   & x & x & 0.63 & 0.73 & \textbf{0.78} & 0.69 & 0.76 & 0.71 & 0.72 \\
   x & x & x &   & 0.71 & 0.81 & 0.75 & 0.72 & 0.75 & \textbf{0.82} & 0.76 \\
   \hdashline
   x & x & x & x & \textbf{0.72} & 0.8  & \textbf{0.78} & \textbf{0.73} & \textbf{0.77} & \textbf{0.82} & \textbf{0.77} \\
   \hline
  \end{tabular}
 \end{center}
\end{table*}

To evaluate the effectiveness of each of the text and network components of CT-TN, we perform a set of ablation of experiments with different sets of these components removed. Table \ref{tab:cttn-vs-ablated} shows the performance scores of the full CT-TN model compared with ablated versions of the model.

We can see that the full CT-TN model achieves top performance in five of the six source-destination target pairs, with the exception of the Sanders-Trump pair where the use of likes only outperforms the full model. For the rest of the target pairs, the full CT-TN either outperforms all ablated models or achieves the same performance as one of the ablated models. Interestingly, however, even if some of the ablated models perform at the same level as the full model on some occasions, there is no consistency on the best ablated model across target pairs. Given the uncertainty on the selection of the best ablated model in each case, it is more reliable to use the full CT-TN model instead, which is more consistent across all target pairs. Indeed, this consistency is also demonstrated in the highest average performance across targets, with an average 0.77 overall. Among the ablated models, those using the likes feature show competitive performance, with the model using only likes and the model combining likes, friends and followers both achieving second-best performance with an average of 0.76. This in turn suggests that, among the network features, the likes are the most useful ones.

\subsection{Reducing the number of shots}

\begin{table*}[htb]
 \normalsize
 \begin{center}
  \caption{Macro-averaged F1 scores on models using different numbers of training shots (100-400) from the destination target.}
  \label{tab:shots}
  \begin{tabular}{|l||llll||llll|}
   \arrayrulecolor{white}\hline\hline\arrayrulecolor{black}
   \cline{2-9}
   \multicolumn{1}{c}{} & \multicolumn{4}{|c||}{Trump $\to$ Sanders} & \multicolumn{4}{c|}{Sanders $\to$ Trump} \\
   \cline{2-9}
   \multicolumn{1}{c|}{} & \#100 & \#200 & \#300 & \#400 & \#100 & \#200 & \#300 & \#400\\
   \hline
   RoBERTa   & 0.24 & 0.28 & 0.49 & 0.53 & 0.31 & 0.35 & 0.54 & 0.59 \\
   CrossNet  & 0.41 & 0.46 & 0.48 & 0.46 & \textbf{0.49} & 0.45 & 0.5  & 0.51 \\
   TGA-Net   & 0.4  & \textbf{0.47} & 0.55 & 0.57 & 0.43 & \textbf{0.48} & 0.58 & 0.6 \\
   \hdashline
   CT-TN     & \textbf{0.44} & 0.4  & \textbf{0.68} & \textbf{0.72} & 0.3  & 0.33 & \textbf{0.78} & \textbf{0.8}  \\
   \hline
   \arrayrulecolor{white}\hline\hline\arrayrulecolor{black}
   
   \cline{2-9}
   \multicolumn{1}{c}{} & \multicolumn{4}{|c||}{Sanders $\to$ Biden} & \multicolumn{4}{c|}{Biden $\to$ Sanders} \\
   \cline{2-9}
   \multicolumn{1}{c|}{} & \#100 & \#200 & \#300 & \#400 & \#100 & \#200 & \#300 & \#400\\
   \hline
   RoBERTa   & \textbf{0.76} & \textbf{0.77} & 0.76 & \textbf{0.78} & 0.5  & 0.47 & 0.58 & 0.66 \\
   CrossNet  & 0.62 & 0.61 & 0.68 & 0.69 & \textbf{0.57} & \textbf{0.56} & 0.58 & 0.58 \\
   TGA-Net   & 0.61 & 0.65 & 0.69 & 0.69 & \textbf{0.57} & \textbf{0.56} & 0.59 & 0.6 \\
   \hdashline
   CT-TN     & 0.59 & 0.73 & \textbf{0.77} & \textbf{0.78} & 0.38 & 0.41 & \textbf{0.73} & \textbf{0.73} \\
   \hline
   \arrayrulecolor{white}\hline\hline\arrayrulecolor{black}
   
   \cline{2-9}
   \multicolumn{1}{c}{} & \multicolumn{4}{|c||}{Trump $\to$ Biden} & \multicolumn{4}{c|}{Biden $\to$ Trump} \\
   \cline{2-9}
   \multicolumn{1}{c|}{} & \#100 & \#200 & \#300 & \#400 & \#100 & \#200 & \#300 & \#400\\
   \hline
   RoBERTa   & \textbf{0.75} & \textbf{0.76} & \textbf{0.78} & \textbf{0.77} & 0.5  & 0.5  & 0.61 & 0.62 \\
   CrossNet  & 0.6  & 0.58 & 0.62 & 0.6  & \textbf{0.52} & \textbf{0.56} & 0.57 & 0.54 \\
   TGA-Net   & 0.67 & 0.68 & 0.69 & 0.69 & 0.49 & 0.5  & 0.56 & 0.59 \\
   \hdashline
   CT-TN     & 0.74 & \textbf{0.76} & 0.77 & \textbf{0.77} & 0.38 & 0.43 & \textbf{0.8}  & \textbf{0.82} \\
   \hline
  \end{tabular}
 \end{center}
\end{table*}

Experiments so far have relied on the use of 400 shots associated with the destination target, showing competitive performance. We are further interested in investigating how CT-TN performs with fewer shots, as well as to assess the number of shots the model needs to outperforms other baseline models.

Table \ref{tab:shots} shows the results for varying numbers of shots, ranging from 100 to 400. These results show a clear trend where the CT-TN model becomes remarkably effective with 300 shots used for training, after which it starts to outperform baseline models, generally by a margin. Conversely, results also show that using 200 or fewer shots is insufficient for CT-TN, where baseline models CrossNet and TGA-Net can perform better. Hence, CT-TN becomes especially reliable as the number of shots increases; however, performance scores are substantially lower for all models when the number of shots is 200 or fewer, hence suggesting that it is worth labelling some more instances up to 300 to achieve a substantial performance gain.

\begin{figure}[htb]
	\centering
		\includegraphics[width=0.5\textwidth]{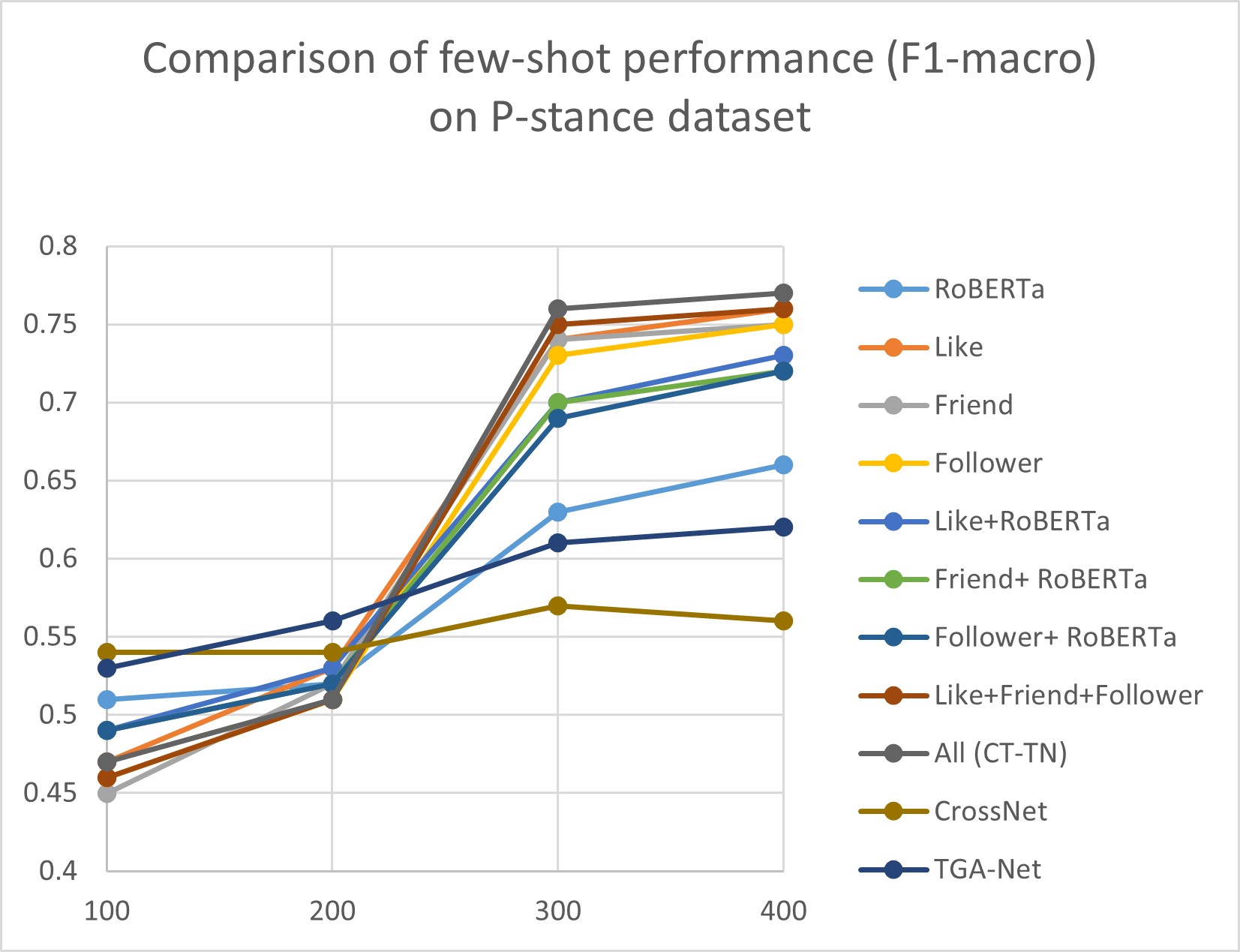}
	\caption{The overall performance of 11 few-shot cross-target tasks.}
	\label{fig:all-plot}
\end{figure}

Figure \ref{fig:all-plot} shows the performance of the full CT-TN model, ablated models as well as baseline models with different numbers of shots used for training. In addition to the results presented in Table \ref{tab:shots}, this figure enables additional visual comparison by also incorporating ablated models. These results reaffirm our previous observations, showing that it is especially after 300 shots that CT-TN and its ablated models become effective. All CT-TN based models achieve a remarkable gain of performance from 200 to 300 shots, which becomes less pronounced when shots are increased from 300 to 400.

\section{Discussion}
\label{sec:Discussion}

Through our experiments, CT-TN has proven to be a very competitive model achieving state-of-the-art performance when it is given a moderate number of training data associated with the destination target. In our case, we have seen that the model can outperform all other baselines with 300 instances pertaining to the destination target. We are however interested in further delving into the performance of CT-TN, which we do next by looking at some of its correct predictions as well as further investigating the structure of the network data it uses.

To better understand the benefits of CT-TN, we delve into some of the examples where CT-TN made a correct prediction and the baseline models made a wrong prediction. We show some of these CT-TN's correct predictions in Table \ref{tab:predictions}. We observe that these are indeed difficult to predict solely from text for an automated model, not least because there are no explicitly positive keywords, often requiring more complex understanding of the text which is not trivial. In situations like these, information derived from the network through CT-TN can be particularly valuable to correct these otherwise challenging predictions.

\begin{table*}[htb]
 \normalsize
 \begin{center}
  \caption{Samples with correct prediction only by CT-TN, where baseline models mispredicted. Examples extracted from experiments for the Biden-Trump target pair.}
  \label{tab:predictions}
  \begin{tabular}{|p{8cm}|l|l|l|l|l|}
   \hline
   \textbf{Tweet}  & \textbf{Real label} & \textbf{CT-TN}        & \textbf{RoBERTa}  & \textbf{CrossNet}  & \textbf{TGA-NET}  \\
   \hline
   Guess we will have to wait forever!! Were working with heartless \#CLOWNS her!! \#TRUMP \#Trump &  FAVOR & FAVOR & AGAINST & AGAINST &AGAINST \\
   \hline
   What was Nancy Pelosi doing when \@realDonaldTrump was putting the \#Coronavirus task force together? Handing out impeachment pens. Voting \#Trump and red down the entire ticket! & FAVOR & FAVOR & AGAINST & AGAINST &AGAINST   \\
   \hline
   People talk about the \#GOP being the party of Lincoln and Reagan, well add \@realDonaldTrump to it because he is a game changing \#POTUS with policy like this! & FAVOR & FAVOR & AGAINST & AGAINST &AGAINST \\
   \hline
  \end{tabular}
 \end{center}
\end{table*}

\begin{figure}[htb]
	\centering
		\includegraphics[width=0.5\textwidth]{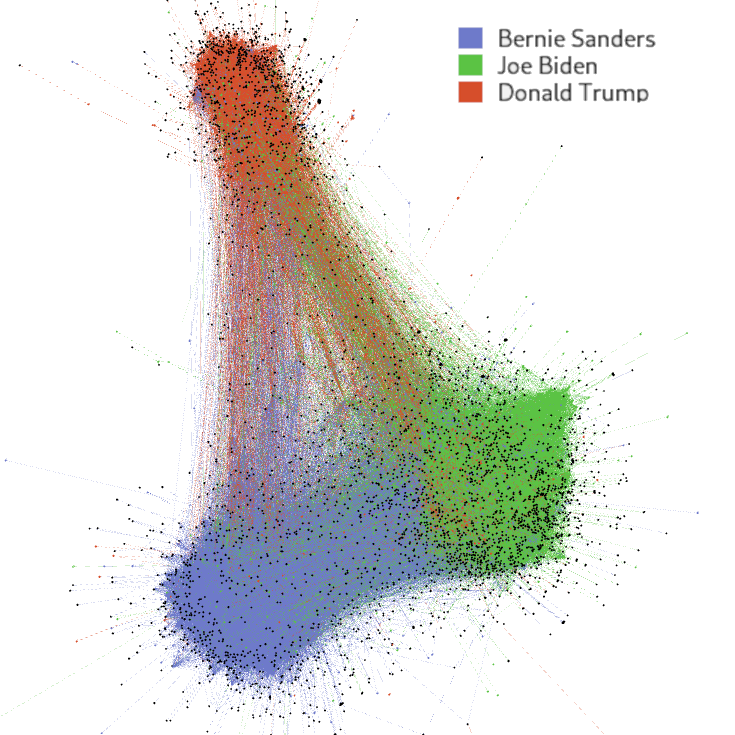}
	\caption{Network visualisation of followers, friends and likes for users expressing supporting stance towards Bernie Sanders (purple), Joe Biden (green) and Donald Trump (red).}
	\label{fig:network}
\end{figure}

Looking at the network data, Figure \ref{fig:network} shows a visualisation of the aggregate of follower, friend and like connections of supporters of each of the political candidates in the dataset, i.e. Bernie Sanders (blue), Joe Biden (green) and Donald Trump (red). Interestingly, we can observe three clear clusters of supporters of each candidate, with strong connections within clusters and fewer connections across clusters. Further, we can also observe that clusters associated with the two candidates of the Democrats, namely Joe Biden and Bernie Sanders, are closer and more strongly connected to each other than any of them is with Republican candidate Donald Trump's cluster. Through our experiments with CT-TN, we demonstrate that, while network information alone would not suffice to achieve top performance on stance detection, it is a valuable feature when used in combination with text, indeed outperforming any ablated models that solely use text or network data.

\section{Conclusion}
\label{sec:conclusion}

To tackle the challenging task cross-target stance detection from social media posts, we have introduced a novel model, CT-TN, which aggregates multimodal text and network embeddings into a model. With a set of experiments across six different source-destination target pairs, we demonstrate the overall effectiveness of CT-TN, outperforming state-of-the-art models such as CrossNet and TGA-Net. While all models struggle with small numbers of shots used for training, CT-TN achieves a noticeable performance gain after 300 shots associated with the destination target are incorporated into the training data. In addition to showing the effectiveness of the novel CT-TN model, we also demonstrate the importance of considering network features for cross-target stance detection, among which the `likes' feature leads to highest performance gains.

Our work in turn opens a set of avenues for future research. While we demonstrate that we can achieve competitive performance with 300+ shots, future work could look into further improving models that perform competitively when fewer shots available, which is particularly important where there are limited resources for labelling new data. Our research demonstrates the effectiveness of CT-TN for 2-class stance detection, while future research could further look into extending it to 3-class stance detection. While datasets enabling cross-target stance detection are very limited to date, not least datasets for which network data can be gathered, we hope to see more suitable datasets in the future, which would also enable further experiments using a bigger set of target pairs.

Our objective with CT-TN has been to demonstrate that the hybrid combination of graph-based embeddings using network interactions and text-based embeddings based on social media posts can lead to an effective approach for the cross-target stance detection task, ourperforming other state-of-the-art text-only models in the task. We do believe however that there is room for further improvement and investigation with other embedding methods, such as wave based architectures for graph representation \cite{matlock2019deep}. Likewise, we chose to develop CT-TN by using a majority voting strategy to combine different embedding inputs, given its superior performance over other methods, such as concatenation of embeddings. We believe that there is room for further exploration with concatenation strategies to try and make them effective for the task, for example by investigating different normalisation methods for the embeddings.

\section*{Acknowledgments}

Parisa Jamadi Khiabani is funded by the Islamic Development Bank (IsDB). We thank the authors of the P-stance dataset for kindly providing us with the tweet IDs which enables us to complement the dataset.

\bibliographystyle{IEEEtran}
\bibliography{cttn}

\end{document}